\documentclass[sn-mathphys,Numbered]{sn-jnl}

\usepackage{graphicx}%
\usepackage{multirow}%
\usepackage{amsmath,amssymb,amsfonts}%
\usepackage{amsthm}%
\usepackage{mathrsfs}%
\usepackage[title]{appendix}%
\usepackage{xcolor}%
\usepackage{textcomp}%
\usepackage{manyfoot}%
\usepackage{booktabs}%
\usepackage{algorithmicx}%
\usepackage{algpseudocode}%
\usepackage{listings}%
\usepackage{bm}%
\usepackage{leftidx}%
\usepackage{tikz}%
\usetikzlibrary{arrows,3d}
\usetikzlibrary{calc}
\usetikzlibrary{math}
\usetikzlibrary{positioning}
\usepackage[ruled]{algorithm2e}
\usepackage{siunitx}
\usetikzlibrary{shapes,arrows,calc,perspective}
\usepackage{pgfplots}
\usepackage{subcaption}

\newcommand{\myFrame}[3]
{
\draw [-latex, red, line width = 0.5mm] (0,0,0) -- (1,0,0);
\draw [-latex, green, line width = 0.5mm] (0,0,0) -- (0,1,0);
\draw [-latex, blue, line width = 0.5mm] (0,0,0) -- (0,0,1);
\shade[ball color = black, opacity = 1] (0,0,0) circle (3pt);
\node [] at #3 {$\mathcal{F}_{#1}$};
\ifnum#2=1
{
\node [red] at (1.25,-0.15,0) {$\leftidx{_{#1}}{x}$};
\node [green] at (0,1.25,0) {$\leftidx{_{#1}}{y}$};
\node [blue] at (0,0,1.25) {$\leftidx{_{#1}}{z}$};
}
\fi
}

\theoremstyle{thmstyleone}%
\theoremstyle{thmstyletwo}%

\theoremstyle{thmstylethree}%

\begin{document}

\parindent0pt

\title[Article Title]{Optimal Dimensioning of Elastic-Link Manipulators regarding Lifetime Estimation}


\author*[1]{\fnm{Klaus} \sur{Zauner}}\email{klaus.zauner@jku.at}

\author[1]{\fnm{Hubert} \sur{Gattringer}}\email{hubert.gattringer@jku.at}

\author[1]{\fnm{Andreas} \sur{Müller}}\email{a.mueller@jku.at}

\affil[1]{\orgdiv{Institute of Robotics}, \orgname{Johannes Kepler University Linz}, \orgaddress{\street{Altenbergerstraße 69}, \city{Linz}, \postcode{4040}, \country{Austria}}}




\abstract{Resourceful operation and design of robots is key for sustainable industrial automation. This will be enabled by lightweight design along with time and energy optimal control of robotic manipulators. Design and control of such systems is intertwined as the control must take into account inherent mechanical compliance while the design must accommodate the dynamic requirements demanded by the control. As basis for such design optimization, a method for estimating the lifetime of elastic link robotic manipulators is presented. This is applied to the geometry optimization of flexible serial manipulators performing pick-and-place operations, where the optimization objective is a combination of overall weight and vibration amplitudes. The lifetime estimation draws from a fatigue analysis combining the rainflow counting algorithm and the method of critical cutting plane. Tresca hypothesis is used to formulate an equivalent stress, and linear damage accumulation is assumed. The final robot geometry is selected from a Pareto front as a tradeoff of lifetime and vibration characteristic. The method is illustrated for a three degrees of freedom articulated robotic manipulator.}




\maketitle

\section{Introduction}\label{introduction}
Currently, lightweight robots are predominantly used in physical human robot collaboration tasks \cite{bischoff2010kuka}. However, the need for cost reduction in manufacturing motivates use of lightweight robots in mass production. Besides lower energy, consumption weight reduction of the structure also leads to material savings, but at the expense of increased elastic vibrations. Highly dynamic pick and place trajectories then inevitably lead to vibrations, which have a negative impact on the performance and can also damage the robot. Appropriate trajectory planning and the control are crucial to counteract those effects \cite{SpringerEtAl}. Approaches taking into account the inherent flexibility of lightweight robots and of robotic manipulators actuated with compliant drives, e.g. serial elastic and variable stiffness actuators, were developed over the last decade \cite{Tsetserukou2008VibrationDC,Malzahn2014,StauferEtAl}.
Vibration damping control requires information about the occuring vibrations, however, which are not available in standard industrial robots. In addition to methods that measure the deformation of the robot arms \cite{Malzahn2014}, some approaches use measurements from acceleration sensors or inertial measurement units (IMUs) \cite{StauferEtAl}. Depending on the method, these measurements have to be processed in a more or less resource-intensive way and implementation on conventional industrial controllers can be quite challenging.

Novel lightweight robots must be designed and dimensioned strategically \cite{met9091020,WANG201948,HermleEberhard2000,PandaEtAl,AlkallaFanni2021,HussainEtAl}, where CAD modeling and subsequent finite element (FEM) simulation are combined \cite{Yin2016}. Instead of computationally expensive FEM simulation and topology optimization, lightweight robots can be constructed from simple geometries, which may later be finetuned. The parameters of the corresponding dynamics model can subsequently be optimized. Every feasible parameter set corresponds to a design candidate. Finding an optimal parameter set under consideration of opposing quality criteria leads to a multi-criterial optimization problem as it often occurs in machine design \cite{MultCritDesign}.

A vital part of a lightweight design is to ensure a certain service lifetime based on damage estimation. A viable approach is the conversion of multi-axial stress states to equivalent uniaxial loads by means of a comparative stress hypothesis in order to calculate damage by counting so-called damage events and comparing them with the material characteristics. These are then accumulated according to a damage hypothesis. The service life after which one has to expect cracks in the structure can be estimated from this damage and the duration of the underlying load cycle \cite{MURAKAMI2005991}.

\begin{figure}[ht]

\begin{tikzpicture}[
declare function={
	theta1(\x)=\x/50*180/pi;
	v1(\x)=-2*0.08*\x*\x;
	vs1(\x)=-2*0.16*\x*180/pi;
	w1(\x)=2*0.04*\x*\x;
	ws1(\x)=2*0.08*\x*180/pi;
	theta2(\x)=\x/50*180/pi;
	v2(\x)=-0.08*\x*\x;
	vs2(\x)=-0.16*\x*180/pi;
	w2(\x)=0.04*\x*\x;
	ws2(\x)=0.08*\x*180/pi;
}
]

\pgfmathsetmacro{\qy}{60};

\draw[line width=0.25mm] (0,0,0) -- (2,0,0);
\myFrame{I}{1}{(-0.75,0,0)};
\coordinate (I) at (0,0,0);
\draw[line width=0.25mm,dashdotted] ($(I)-(0,0.5,0)$) -- ($(I)+(0,0.5,0)$);
\begin{scope}[canvas is xz plane at y=0]
\draw[-latex,line width=0.3mm] (1.8,0,0) arc (0:-\qy:1.8);
\end{scope}

\begin{scope}[shift={(0,0,0)},rotate around y=\qy]
\coordinate (q2) at (0,2,0);
\draw[line width=0.25mm] (I) -- ($(q2)+(0,1.25,0)$);
\draw[line width=1mm] (I) -- (q2);
\draw[line width=0.25mm] (I) -- (2,0,0);
\myFrame{1}{0}{(-0.75,0,0)};
\end{scope}

\begin{scope}[shift={(q2)},rotate around y=\qy,rotate around z=45]
\draw[line width=0.25mm,dashdotted] ($(q2)-(0,0,1)$) -- ($(q2)+(0,0,1)$);
\coordinate (q3) at (3,0,0);
\draw[line width=0.25mm] (q2) -- ($(q3)+(1.25,0,0)$);
\draw[line width=1mm] (q2) -- (q3);
\draw[line width = 1mm, scale=1, domain=0:3, smooth, variable=\x] plot ({\x},{v1(\x)},{w1(\x)});
\coordinate (q3') at (3,{v1(3)},{w1(3)});
\myFrame{2}{0}{(-0.75,0,0)};
\draw[-latex,line width=0.3mm] (1,0,0) arc (0:45:1);
\begin{scope}[shift={(3,{v1(3)},{w1(3)})},rotate around x=theta1(3),rotate around y=-ws1(3),rotate around z=vs1(3)]
\draw[line width=0.25mm] (0,0,0) -- (1.25,0,0);
\begin{scope}[rotate around z=-10]
\draw[line width = 1mm, scale=1, domain=0:3, smooth, variable=\x] plot ({\x},{v2(\x)},{w2(\x)});
\draw[line width=0.25mm,dashdotted] ($(q3')-(0,0,1)$) -- ($(q3')+(0,0,1)$);
\myFrame{3'}{0}{(-0.75,0,0)};
\begin{scope}[shift={(3,{v2(3)},{w2(3)})},rotate around x=theta2(3),rotate around y=-ws2(3),rotate around z=vs2(3)]
\myFrame{EE'}{0}{(-1,0,1.5)};
\coordinate (EE') at (0,0,0);
\end{scope}
\end{scope}
\end{scope}
\end{scope}

\begin{scope}[shift={(q3)},rotate around y=\qy,rotate around z=-10]
\draw[line width=0.25mm,dashdotted] ($(q3)-(0,0,1)$) -- ($(q3)+(0,0,1)$);
\coordinate (EE) at (3,0,0);
\draw[line width=1mm] (q3) -- (EE);
\myFrame{3}{0}{(-0.75,0,0)};
\draw[-latex,line width=0.3mm] (1,0,0) arc (0:55:1);
\begin{scope}[shift={(EE)}]
\myFrame{EE}{0}{(-0.75,0.5,0)};
\end{scope}
\end{scope}

\draw[-latex,line width=0.5mm] (EE) -- node[midway,right] {$\Delta r_{EE}$} (EE');

\node[] at (9,3,0) {\includegraphics[width=0.35\textwidth]{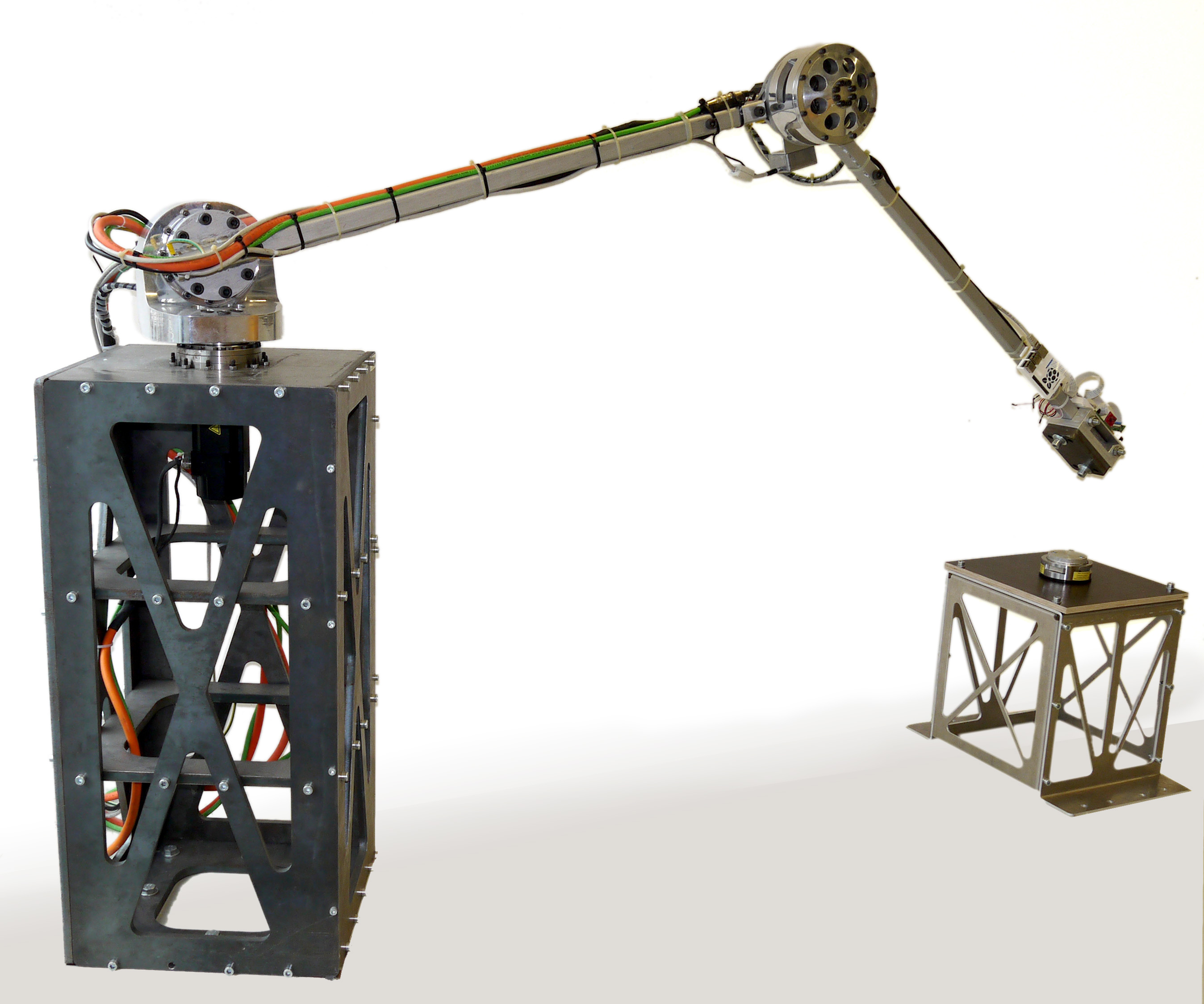}};
\end{tikzpicture}

\caption{Elastic manipulator ELLA, Institute of Robotics @ JKU}
\label{fig:ELLA}
\end{figure}
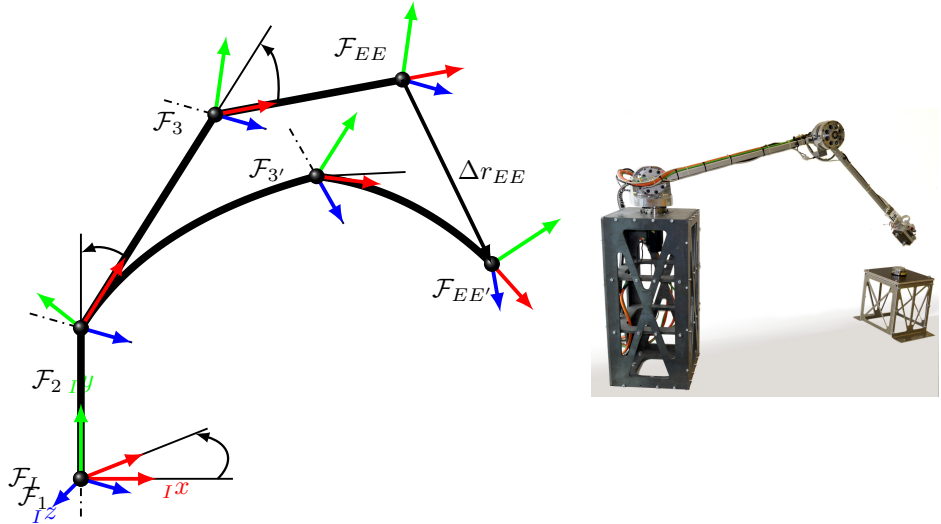

The contribution of this paper is a method for fatigue life estimation of robotic manipulators as depicted in Fig. \ref{fig:ELLA}, and its use in design optimization. It builds upon preliminary work presented in \cite{ECCOMAS}. The proposed method is a combination of the critical cutting plane method \cite{Fatemi2011,MOREL199987} and the linear damage accumulation according to Palmgren and Miner \cite{Miner}. The rainflow counting algorithm \cite{Matsuishi1968,ASTM,Anthes1997ModifiedRC} is used to identify the load cycles responsible for the damage.

The paper is organized as follows. Modeling and the introduction of the design parameters is carried out in section \ref{sec:modeling}. Section \ref{sec:loadCases} covers the generation of simulation data. The definition and evaluation of design candidates as well as the lifetime estimation is carried out in sections \ref{sec:candidateDesigns} and \ref{sec:fatigue}. The methodology, consisting of a parameter study and subsequent fatigue life estimation is illustrated in section \ref{applicationAndResults} using the example of an elastic 3DOF articulated arm robot. Section \ref{sec:conclusion} concludes the paper and gives an outlook on possible future research topics.

\section{Modeling}\label{sec:modeling}
The modeling of elastic robots can be carried out in variable detail with regard to the elasticities taken into account. In addition to the gear elasticities and the elasticities of the flexible links, those of joint structures and bearings can also be taken into account. In this paper the focus is on the durability estimation of the elastic links of a serial robot. Detailing on the modeling, especially on the elastic beam modeling for small deformations and the derivation of the equations of motion, can be found in \cite{Bremer_08} for example.

\subsection{Multi-Elastic Links}
In general, the links of industrial robots have quite complex geometries. The calculation of their deformations under load therefore usually requires the use of FEM simulations. In the following, the links of a lightweight manipulators are regarded as Euler-Bernoulli beam, which are
\begin{itemize}
\item the beam's cross-sections stay normal to the beam's axis,
\item the beam's cross-sections stay planar,
\item the beam's deflections are small compared to its longitudinal expansion,
\item the beam consists of isotropic material and follows Hooke's law.
\end{itemize}

To derive the equation of motion of the elastic beam, it is divided into infinitesimal slices, hereinafter referred to as elements and each with its own frame $\mathcal{F}_E$, see Fig. \ref{fig:subsysII}. They can be imagined as being threaded along the axis of the beam, which runs through the center of mass (COM) of each of the elements. The beam's longitudinal axis is assigned the coordinate $\xi\in[0,L]$, with $L$ the length of the undeformed beam. The reference frame $\mathcal{F}_B$ of the undeformed configuration is at the origin $\xi=0$.

\newcommand{\savedx}{0}
\newcommand{\savedy}{0}
\newcommand{\savedz}{0}

\newcommand{\crosssection}%
{  \coordinate (a) at (0,0.5,0.25);
    \coordinate (b) at (0,0.5,-0.25);
    \coordinate (c) at (0,-0.5,-0.25);
    \coordinate (d) at (0,-0.5,0.25);
    \draw [line width=0.5] (a)--(b) (b)--(c) (c)--(d) (d)--(a);
}

\newcommand{\crosssectionInner}%
{  \coordinate (a) at (0,0.4,0.15);
    \coordinate (b) at (0,0.4,-0.15);
    \coordinate (c) at (0,-0.4,-0.15);
    \coordinate (d) at (0,-0.4,0.15);
    \draw [line width=0.5] (a)--(b) (b)--(c) (c)--(d) (d)--(a);
}

\newcommand{\rotateRPY}[4][0/0/0]
{  
    \pgfmathsetmacro{\rollangle}{#2}
    \pgfmathsetmacro{\pitchangle}{#3}
    \pgfmathsetmacro{\yawangle}{#4}

    \pgfmathsetmacro{\newxx}{cos(\yawangle)*cos(\pitchangle)}
    \pgfmathsetmacro{\newxy}{sin(\yawangle)*cos(\pitchangle)}
    \pgfmathsetmacro{\newxz}{-sin(\pitchangle)}
    \path (\newxx,\newxy,\newxz);
    \pgfgetlastxy{\nxx}{\nxy};

    \pgfmathsetmacro{\newyx}{cos(\yawangle)*sin(\pitchangle)*sin(\rollangle)-sin(\yawangle)*cos(\rollangle)}
    \pgfmathsetmacro{\newyy}{sin(\yawangle)*sin(\pitchangle)*sin(\rollangle)+ cos(\yawangle)*cos(\rollangle)}
    \pgfmathsetmacro{\newyz}{cos(\pitchangle)*sin(\rollangle)}
    \path (\newyx,\newyy,\newyz);
    \pgfgetlastxy{\nyx}{\nyy};

    \pgfmathsetmacro{\newzx}{cos(\yawangle)*sin(\pitchangle)*cos(\rollangle)+ sin(\yawangle)*sin(\rollangle)}
    \pgfmathsetmacro{\newzy}{sin(\yawangle)*sin(\pitchangle)*cos(\rollangle)-cos(\yawangle)*sin(\rollangle)}
    \pgfmathsetmacro{\newzz}{cos(\pitchangle)*cos(\rollangle)}
    \path (\newzx,\newzy,\newzz);
    \pgfgetlastxy{\nzx}{\nzy};

    \foreach \x/\y/\z in {#1}
    {  \pgfmathsetmacro{\transformedx}{\x*\newxx+\y*\newyx+\z*\newzx}
        \pgfmathsetmacro{\transformedy}{\x*\newxy+\y*\newyy+\z*\newzy}
        \pgfmathsetmacro{\transformedz}{\x*\newxz+\y*\newyz+\z*\newzz}
        \xdef\savedx{\transformedx}
        \xdef\savedy{\transformedy}
        \xdef\savedz{\transformedz}     
    }
}

\tikzset{RPY/.style={x={(\nxx,\nxy)},y={(\nyx,\nyy)},z={(\nzx,\nzy)}}}

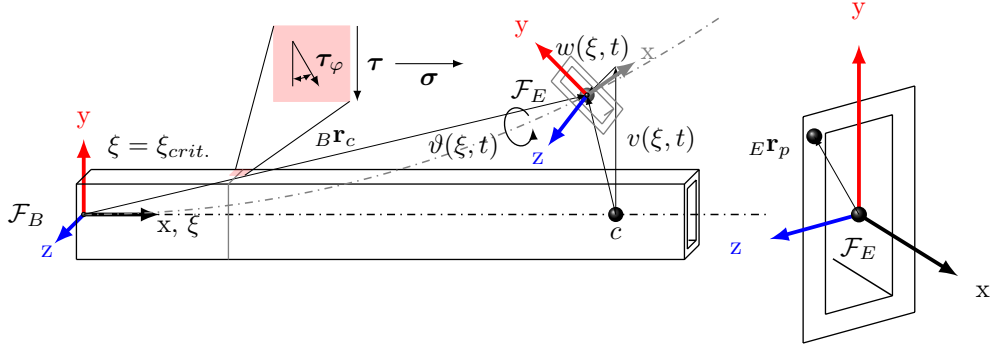
\begin{figure}[ht]
\begin{center}
\begin{tikzpicture}[
declare function={
	theta(\x)=\x/50*180/pi;
	v(\x)=0.04*\x*\x;
	vs(\x)=0.08*\x*180/pi;
	w(\x)=0.02*\x*\x;
	ws(\x)=0.04*\x*180/pi;
}
]

\draw [-latex, line width = 0.5mm] (0,0,0) -- (1,0,0);
\node [] at (1.25,-0.15,0) {x, $\xi$};
\draw [-latex, red, line width = 0.5mm] (0,0,0) -- (0,1,0);
\node [red] at (0,1.25,0) {y};
\draw [-latex, blue, line width = 0.5mm] (0,0,0) -- (0,0,1);
\node [blue] at (0,0,1.25) {z};
\shade[ball color = black, opacity = 1] (0,0,0) circle (1pt);
\node [] at (-0.75,0,0) {$\mathcal{F}_B$};

\draw [line width=0.5] (0,-0.5,0.25) -- (0,0.5,0.25) -- (0,0.5,-0.25);
\draw [line width=0.5] (0,-0.5,0.25) -- (8,-0.5,0.25);
\draw [line width=0.5] (0,0.5,0.25) --  (8,0.5,0.25);
\draw [line width=0.5] (0,0.5,-0.25) -- (8,0.5,-0.25);

\begin{scope}[shift={(8,0,0)}]
\crosssection
\crosssectionInner
\draw [line width = 0.2mm] (0,-0.4,-0.15) -- (-0.1,-0.4,-0.15);
\end{scope}

\draw [line width = 0.2mm, dashdotted] (0,0,0) -- (9,0,0);
\draw[line width = 0.2mm, dashdotted, scale=1, domain=0:9, smooth, variable=\x, gray] plot ({\x},{v(\x)} ,{w(\x)});

\rotateRPY{theta(6)}{-ws(6)}{vs(6)}
\begin{scope}[draw=gray, text=gray,fill=gray,shift={(6,{v(6)},{w(6)})}, RPY]
\begin{scope}[canvas is yz plane at x=0]
\draw [-latex, line width=0.5, black] (0,-0.25) arc (-90:230:0.25);
\end{scope}
\end{scope}

\shade[ball color = black, opacity = 1] (7,0,0) circle (1mm);
\shade[ball color = gray, opacity = 1] (7,{v(7)},{w(7)}) circle (1mm);
\draw [-latex, line width = 0.25] (7,0,0) -- (7,{v(7)},0);
\draw [-latex, line width = 0.25] (7,{v(7)},0) -- (7,{v(7)},{w(7)});
\draw [-latex, line width = 0.25] (7,0,0) -- ((7,{v(7)},{w(7)});
\draw [-latex, line width = 0.25] (0,0,0) -- (7,{v(7)},{w(7)}) node [midway, above] {$\leftidx{_B}{\mathbf{r}}{_c}$};
\rotateRPY{theta(8)}{-ws(8)}{vs(8)}
\begin{scope}[draw=gray, text=gray,fill=gray,shift={(7,{v(7},{w(7)})}, RPY]
\crosssection
\crosssectionInner
\draw [line width = 0.2mm] (0,-0.4,-0.15) -- (-0.2,-0.4,-0.15);
\draw [-latex, line width = 0.5mm] (0,0,0) -- (1,0,0);
\node [] at (1.25,0,0) {x};
\draw [-latex, red, line width = 0.5mm] (0,0,0) -- (0,1,0);
\node [red] at (0,1.25,0) {y};
\draw [-latex, blue, line width = 0.5mm] (0,0,0) -- (0,0,1);
\node [blue] at (0,0,1.25) {z};
\shade[ball color = black, opacity = 1] (0,0,0) circle (1pt);
\node [black] at (0,0.65,0.6) {$\mathcal{F}_E$};
\end{scope}

\rotateRPY{0}{-45}{0}
\begin{scope}[scale=3,shift={(3.4,0,0)}, RPY]
\crosssection
\crosssectionInner
\draw [line width = 0.2mm] (0,-0.4,-0.15) -- (-0.6,-0.4,-0.15);
\draw [-latex, line width = 0.5mm] (0,0,0) -- (1,0,0);
\node [] at (1.25,0,0) {x};
\draw [-latex, red, line width = 0.5mm] (0,0,0) -- (0,0.75,0);
\node [red] at (0,0.9,0) {y};
\draw [-latex, blue, line width = 0.5mm] (0,0,0) -- (0,0,0.4);
\node [blue] at (0,0,0.55) {z};
\shade[ball color = black, opacity = 1] (0,0,0) circle (1pt);
\node [black] at (0,-0.15,0) {$\mathcal{F}_E$};
\shade[ball color = black, opacity = 1] (0,0.4,0.2) circle (1pt);
\draw [-latex, line width = 0.25] (0,0,0) -- (0,0.4,0.2); 
\node [] at (0,0.4,0.4) {$\leftidx{_E}{\mathbf{r}}{_p}$};
\end{scope}

\begin{scope}[canvas is xy plane at z=0]
\node [] at (7.6,1) {$v(\xi,t)$};
\node [] at (6.7,2.2) {$w(\xi,t)$};
\node [] at (1,0.9) {$\xi=\xi_{crit.}$};
\node [] at (5,0.9) {$\vartheta(\xi,t)$};
\node [] at (7,-0.25) {$c$};

\draw [red, fill=red, opacity = 0.2] (2.5cm,1.5cm) rectangle ++(1cm,1cm);
\coordinate (h1) at (2.5cm,2.5cm);
\coordinate (h2) at (3.5cm,1.5cm);
\draw [-latex] (3.6,2.5) -- (3.6,1.5) node [midway, right] {$\boldsymbol{\tau}$};
\draw [-latex] (4.1,2) -- (5,2) node [midway, below] {$\bm{\sigma}$};
\draw [] (2.75,1.7) -- (2.75,2.3);
\draw [-latex] (2.75,2.3) -- ++(300:0.7) node [midway, right] {$\bm{\tau}_{\varphi}$};
\draw[latex-latex, line width = 0.1] (2.75,1.8) arc (270:300:0.5);
\end{scope}



\begin{scope}[draw=gray, text=gray,fill=gray,shift={(2,0,0)}]
\draw [line width=0.5] (0,-0.5,0.25) -- (0,0.5,0.25) -- (0,0.5,-0.25);
\begin{scope}[canvas is xz plane at y=0.5]
\draw [red, fill=red, opacity = 0.2] (-0.1,-0.05) -- (0.1,-0.05) -- (0.1,-0.25) -- (-0.1,-0.25) -- (-0.1,-0.05);
\coordinate (h3) at ((0.1,-0.05);
\coordinate (h4) at (-0.1,-0.25);
\end{scope}
\end{scope}

\draw [] (h1) -- (h4);
\draw [] (h2) -- (h3);

\end{tikzpicture}
\caption{Elastic link (beam) of an articulated robot, with elastic displacements and critical point for the subsequent lifetime estimation}
\label{fig:subsysII}
\end{center}
\end{figure}

The COM of an element in the undeformed configuration is denoted as $\leftidx{_B}{\mathbf{r}}{_c^T}=(\xi,0,0)$. In the deformed configuration, the vector to the COM and the $\mathcal{F}_E$ frame's  orientation are given to
\begin{equation}
\leftidx{_B}{\mathbf{r}}{_c^T}=
\begin{pmatrix}
\xi & v(\xi,t) & w(\xi,t)
\end{pmatrix},\,
\leftidx{_B}{\bm{\varphi}}{_c^T}=
\begin{pmatrix}
\vartheta(\xi,t) & -w'(\xi,t) & v'(\xi,t)
\end{pmatrix},
\end {equation}
with $v(\xi,t)$ and $w(\xi,t)$ denoting the time dependent displacement along the $y$-axis and $z$-axis respectively. The torsional angle around the $x$-axis is $\vartheta(\xi,t)$. The normally partial equation of motion of the beam is decoupled with regard to spatial and temporal dependencies, $v(\xi,t)=\mathbf{v}(\xi)^T\mathbf{q}_v(t),\,w(\xi,t)=\mathbf{w}(\xi)^T\mathbf{q}_w(t)$ and $\vartheta(\xi,t)=\bm{\vartheta}(\xi)^T\mathbf{q}_{\vartheta}(t)$, using the direct Ritz-method. Vectors $\mathbf{v}(\xi)$, $\mathbf{w}(\xi)$ and $\bm{\vartheta}(\xi)$ denote the shape functions. Minimal coordinates $\mathbf{q}_e^T=\left(\mathbf{q}_v^T\,\mathbf{q}_w^T\,\mathbf{q}_{\vartheta}^T\right)$ are the time dependent Ritz coefficients.

The inertia parameters of the differential beam element are given by the differential mass
\begin{equation}
dm_E=\rho A_Bd\xi,
\end{equation}
and the inertia tensor
\begin{equation}
d\mathbf{J}_E=
\begin{bmatrix}
\rho I_D & 0 & 0\\
0 & \rho I_y & 0\\
0 & 0 & \rho I_z
\end{bmatrix}d\xi,
\end{equation}
resolved in the element fixed frame $\mathcal{F}_E$, which is a principal axis system for beams with symmetric cross sections. The cross-sectional area is denoted by $A_B$, $I_y$ and $I_z$ are the area moments of inertia and $I_D$ is the torsional moment of inertia.

As the beam cross-section remains flat and normal to the beam axis in the deformed state, the position vector of any material point of an element at $\xi=\xi_{\text{crit}}$ ($\xi_{\text{crit}}=0$ for one side clamped beam)  is
\begin{equation}
\label{eq:materialPoint}
\leftidx{_E}{\mathbf{r}}{_p^T}=(0\,y\,z).
\end{equation}
Linear elastic material behavior
\begin{equation}
\label{eq:multiaxialHooke}
\bm{\sigma}=\mathbf{H}\bm{\varepsilon}
\end{equation}
with the Hooke matrix $\mathbf{H}$ and the strain vector $\bm{\varepsilon}^T=\left((-v''y-w''z)\,0\,0\,-\vartheta'z\,0\,\vartheta'y\right)$ is assumed. Taking into account the assumptions of the Euler-Bernoulli beam, the strain vector can be rewritten in a reduced form, including only non-zero elements, as
\begin{equation}
\bm{\varepsilon}_{\text{red}}=
\begin{pmatrix}
\varepsilon_1\\
\varepsilon_2\\
\varepsilon_3
\end{pmatrix}=
\begin{bmatrix}
0 & -y & -z\\
-z & 0 & 0\\
y & 0 & 0\\
\end{bmatrix}
\underset{\bm{\kappa}}{\underbrace{
\begin{pmatrix}
\vartheta'\\
v''\\
w''
\end{pmatrix}}}
\end{equation}
with the curvature $\bm{\kappa}$. The corresponding, reduced $\mathbf{H}$-matrix is
\begin{equation}
\mathbf{H}_{\text{red}}=
\begin{bmatrix}
E & 0 & 0\\
0 & G & 0\\
0 & 0 & G
\end{bmatrix}
\end{equation}
with the shear modulus $G=\frac{E}{2(1+\nu)}$, where $E$ is the Young's modulus and $\nu$ denotes Poisson's ratio.
The beam returns to its undeformed state without an external load. This behavior is modeled by the reaction force $\mathbf{Q}_{el}=-\mathbf{K}_{el}\mathbf{q}_e$ acting on the elastic coordinates, where the stiffness matrix is
\begin{equation}
\label{eq:stiffnessMatrix}
\mathbf{K}_{el}=\int\limits_0^L
\begin{bmatrix}
GI_D\bm{\vartheta}''\bm{\vartheta}''^T & 0 & 0\\
0 & EI_z\mathbf{v}''\mathbf{v}''^T & 0\\
0 & 0 & EI_y\mathbf{w}''\mathbf{w}''^T
\end{bmatrix}d\xi.
\end{equation}
It is derived by calculating the deformation related potential energy of the beam.



Dynamics modeling according to \cite{Bremer_08} yields the equation of motion (EOM)
\begin{equation}
\label{eq:EOMs}
\mathbf{M}(\mathbf{q},\mathbf{p})\ddot{\mathbf{q}}+\mathbf{G}(\mathbf{q},\dot{\mathbf{q}},\mathbf{p})\dot{\mathbf{q}}+\mathbf{g}(\mathbf{q},\mathbf{p})=\mathbf{Q}_{\text{M}},
\end{equation}
with the generalized mass matrix $\mathbf{M}(\mathbf{q},\mathbf{p})$, the vector of generalized Coriolis and centrifugal forces $\mathbf{G}(\mathbf{q},\dot{\mathbf{q}},\mathbf{p})\dot{\mathbf{q}}$, gravitational forces $\mathbf{g}(\mathbf{q},\mathbf{p})$ and the motor torques $\mathbf{Q}_{\text{M}}$. The generalized coordinates $\mathbf{q}$ consist of the actuated motor coordinates $\mathbf{q}_M$, the link sided coordinates $\mathbf{q}_L$ (in case of elastic gearboxes) and the elastic coordinates $\mathbf{q}_e$. Length-, geometry- and material-parameters are summarized in $\mathbf{p}$.

\section{Load-Case Simulation}\label{sec:loadCases}
The desired model-based design requires data on vibrations and occurring stresses. These must be simulated for a relevant trajectory. A standard application of a lightweight robots are pick and place tasks, ideally with a minimal task duration $t_{\text{task}}$. In general, except when possible collision objects have to be taken into account, only the position $\leftidx{_I}{\mathbf{r}}{_E}$ and the orientation $\leftidx{_I}{\bm{\varphi}}{_E}$ (e.\,g. in Cardan-angles) of the end effector, resolved in the inertial frame $\mathcal{F}_I$, are of particular interest when planning robot motions. The desired time optimization of the movement makes it necessary to consider speed, acceleration and torque limits. Therefore the joint trajectories $\mathbf{q}_d$ need to be computed by the inverse kinematics.\\



Simulating the robot dynamics at a desired motion $\mathbf{q}_d$ requires solving (\ref{eq:EOMs}), in combination with an industrial cascaded position controller, for $\ddot{\mathbf{q}}$. 
An efficient $\mathcal{O}(n)$-algorithm for multibody systems in subsystem representation, as stated in \cite{Bremer_08}, is used to this end. 
Numerical integration is pursued with an appropriate solver for numerically stiff problems (e.\,g. ode15s).

\section{Determination of Candidate Designs}\label{sec:candidateDesigns}
The first step is to identify design candidates that meet the primary quality criteria of weight reduction and low vibration tendency. The structure of the considered manipulator shall be fixed. Therefore its dynamic behavior is determined by EOMs of the form (\ref{eq:EOMs}) with parameters $\mathbf{p}$. Every candidate design is basically a set $\mathbf{p}_i\in\mathbf{P}$ out of the set of permissible parameters $\mathbf{P}$. Vector $\mathbf{p}$ contains several parameters that  determine the workspace, which is usually a specification for the robot. Therefore $\mathbf{p}\in\mathbb{R}^{n_p}$ is partitioned into fixed parameters $\mathbf{p}_f\in\mathbb{R}^{n_f}$  and parameters $\mathbf{p}_v\in\mathbb{R}^{n_v}$, which can be adjusted to influence the dynamic behavior of the manipulator.\\

From the constructive point of view, the parameters that are easiest to change are those of the elastic links. As emphasized in the modeling section \ref{sec:modeling}, these are the material-dependent parameters density and Young's modulus, as well as the geometric parameters cross-sectional area and area moment of inertia. The latter are defined by profile shapes and their dimensions. These can be varied in certain ranges, whereby constructive requirements must also be taken into account. However, these considerations are beyond the scope of this paper and are only carried out for the specific example in section \ref{applicationAndResults}.\\

The aforementioned goals give rise to the multicriterial optimization \cite{MultCritDesign} problem
\begin{equation}
\underset{\mathbf{p}\in\mathbf{P}}{\text{min}}\,\left(J_{\text{m}}\left(\mathbf{p}\right),J_{\text{vib.}}\left(\mathbf{p},\mathbf{q}_d(t)\right)\right)
\end{equation}
with the $m=2$ quality criteria $J_{\text{m}}\left(\mathbf{p}\right)$ which assesses the design dependent robot weight compared to an initial design and $J_{\text{vib}}\left(\mathbf{p},\mathbf{q}_d(t)\right)$ which denotes a quality criteria regarding the endeffector vibration, e.\,g. the maximum derivation from the desired position during a critical phase of the trajectory.\\

As these quality criteria are opponing each other, there exists no design that can be clearly identified as the best solution. The so called Pareto front $\mathbf{F}_P$, see Fig. \ref{fig:paretoDiag}, is defined as the set of candidates that are not outperformed by any other one.

The configurations of the Pareto front are the candidates for the fatigue analysis.


\section{Fatigue Analysis}\label{sec:fatigue}
The use of lightweight robots for repetitive tasks to be performed as quickly as possible leads to recurring loads on the bearings of the moving parts (joints, linear guides, ...) and the structural elements (joint structures, arms, ...). Lightweight design is a decision between the lightest possible constructions and the necessary durability. As this does not allow for large reserves with regard to dynamic loads, it must be estimated during the design process whether the manipulator will achieve the required service life assuming representative load cases. Any correspondingly large load will cause further damage to the structure, so there must be a number of repetitions after which one must at least expect a part of the structure to fail. Let $D\in[0,1]$ denote a quantification of the damage at an interesting point of the structure during one execution of the representative task. With the task duration $t_{\text{task}}$, an estimate for the lifetime $t_{\text{life}}$ (of exactly this part of the structure) is given by
\begin{equation}
\label{eq:lifetimeEstimate}
t_{\text{life}}=\frac{t_{\text{task}}}{D}.
\end{equation}

The damage increases over time, therefore an incremental growth $\Delta D$ is introduced, which is related to so called damage events.\\

In the following, the necessary steps for calculating the lifetime estimate are shown.

\subsection{Cutting Plane Dependent Stress}
\begin{figure}[ht]
\begin{center}
\begin{tikzpicture}[3d view = {25}{15},
declare function={
	theta(\x)=\x/50*180/pi;
	v(\x)=0.04*\x*\x;
	vs(\x)=0.08*\x*180/pi;
	w(\x)=0.02*\x*\x;
	ws(\x)=0.04*\x*180/pi;
}
]

\begin{scope}[canvas is yz plane at x = 0]
        
\draw[fill = lightgray] (-2, -2) rectangle (2, 2);
\draw[fill = white] (-1.8, -1.8) rectangle (1.8, 1.8);
        
\end{scope}  

\draw [] (0,-2,-2) -- (-3,-2,-2);
\draw [] (0,-2,2) -- (-3,-2,2);  
\draw [] (0,2,2) -- (-3,2,2);  
\draw [] (0,1.8,-1.8) -- (-1.5,1.8,-1.8);  

\coordinate (I) at (0,1,1.9);
\filldraw (I) circle (2pt);
\draw[-latex, line width = 0.5mm] ($(I)+(0,0,0.5)$) -- ($(I)+(0,0,-1)$);
\node[] at ($(I)+(0,0,-1.25)$) {$z$};
\draw[-latex, line width = 0.5mm] (I) -- ($(I)+(0,-1,0)$);
\node[] at ($(I)+(0,-1.25,0)$) {$y$};
\draw[-latex, line width = 0.5mm] (I) -- ($(I)+(1,0,0)$);
\node[] at ($(I)+(1.25,0,0)$) {$x$};

\begin{scope}[shift={(0,1,2.4)}]
\draw [line width=0.5] (0,0) -- (0.5,0);
\draw [-latex, line width=0.5] (0.4,0) arc (0:-180:0.4);
\end{scope}

\node [] at (0,0,3) {$\varphi$};

\rotateRPY{0}{0}{-25}
\begin{scope}[draw=gray, text=gray,fill=gray,shift={(0,1,1.9)}, RPY]
\draw [-latex, red, line width = 0.5mm] (0,0,0) -- (0,-1,0);
\node [red] at (0,-1.25,0) {n};
\draw [-latex, blue, line width = 0.5mm] (0,0,0) -- (1,0,0);
\node [blue] at (1.25,0,0) {m};
\shade[ball color = black, opacity = 1] (0,0,0) circle (2pt);
\end{scope}

\draw [dashdotted] (-0.8,0,0) -- (4,0,0);
\draw[line width = 0.2mm, dashdotted, scale=1, domain=0:4, smooth, variable=\x, gray] plot ({\x},{v(\x)} ,{w(\x)});
\node [] at (2,2,0) {curvature $\kappa$};

\rotateRPY{180}{0}{0}
\begin{scope}[draw=gray, text=gray,fill=gray,shift={(0,0,0)}, RPY]
\draw [-latex, line width = 0.5mm] (0,0,0) -- (1,0,0);
\node [] at (1.25,0,0.3) {$\leftidx{_E}{x}{}$};
\draw [-latex, red, line width = 0.5mm] (0,0,0) -- (0,1,0);
\node [red] at (0,1.25,0.2) {$\leftidx{_E}{y}{}$};
\draw [-latex, blue, line width = 0.5mm] (0,0,0) -- (0,0,1);
\node [blue] at (0.3,0,1) {$\leftidx{_E}{z}{}$};
\shade[ball color = black, opacity = 1] (0,0,0) circle (2pt);
\end{scope}

\begin{scope}[canvas is yz plane at x = 0.5]
\draw [-latex, line width=0.5] (0,-0.25) arc (-90:230:0.25);
\end{scope}

\begin{scope}[canvas is xy plane at z = 0]
\draw [-latex, line width=0.5] (65:0.5) arc (65:-65:0.5);
\end{scope}

\begin{scope}[canvas is xz plane at y = 0]
\draw [-latex, line width=0.5] (-65:0.5) arc (-65:65:0.5);
\end{scope}

\coordinate (OE) at (5,3,-2);
\begin{scope}[canvas is xy plane at z = 3, shift = {(5,3)}]
\draw[fill = lightgray] (-1, -1) rectangle (1,1);
\end{scope}

\begin{scope}[canvas is xz plane at y = 2, shift = {(5,2)}]
\draw[fill = lightgray] (-1, -1) rectangle (1,1);
\end{scope}

\begin{scope}[canvas is yz plane at x = 6, shift = {(3,2)}]
\draw[fill = lightgray] (-1, -1) rectangle (1,1);
\end{scope}

\coordinate (yPlane) at (5,2,2);
\draw [-latex, line width = 0.5mm] (yPlane) -- ++(0.75,0,0);
\node [] at ($(yPlane)+(0.5,0,-0.25)$) {$\sigma_{xy}$};
\draw [-latex, line width = 0.5mm] (yPlane) -- ++(0,-0.75,0);
\node [] at ($(yPlane)+(0.1,-1,-0.2)$) {$\sigma_{yy}$};
\draw [-latex, line width = 0.5mm] (yPlane) -- ++(0,0,0.75);
\node [] at ($(yPlane)+(-0.6,0,0.5)$) {$\sigma_{yz}$};
\shade[ball color = black, opacity = 1] (yPlane) circle (2pt);

\coordinate (zPlane) at (5,3,3);
\draw [-latex, line width = 0.5mm] (zPlane) -- ++(0.75,0,0);
\node [] at ($(zPlane)+(1,0,0)$) {$\sigma_{xz}$};
\draw [-latex, line width = 0.5mm] (zPlane) -- ++(0,-0.75,0);
\node [] at ($(zPlane)+(-0.5,-0.5,0)$) {$\sigma_{yz}$};
\draw [-latex, line width = 0.5mm] (zPlane) -- ++(0,0,0.75);
\node [] at ($(zPlane)+(0.3,0,1)$) {$\sigma_{zz}$};
\shade[ball color = black, opacity = 1] (zPlane) circle (2pt);

\coordinate (xPlane) at (6,3,2);
\draw [-latex, line width = 0.5mm] (xPlane) -- ++(0.75,0,0);
\node [] at ($(xPlane)+(1,0,-0.25)$) {$\sigma_{xx}$};
\draw [-latex, line width = 0.5mm] (xPlane) -- ++(0,-0.75,0);
\node [] at ($(xPlane)+(0.5,-1,-0.1)$) {$\sigma_{xy}$};
\draw [-latex, line width = 0.5mm] (xPlane) -- ++(0,0,0.75);
\node [] at ($(xPlane)+(0,0.75,0.5)$) {$\sigma_{xz}$};
\shade[ball color = black, opacity = 1] (xPlane) circle (2pt);

\coordinate (OE) at (5,3,2);
\shade[ball color = black, opacity = 1] (OE) circle (2pt);
\draw [-latex, line width = 0.5mm] (0,1,1.9) -- (5,3,2);

\end{tikzpicture}
\caption{Stress state in a thin walled beam}
\label{fig:stressStateBeam}
\end{center}
\end{figure}
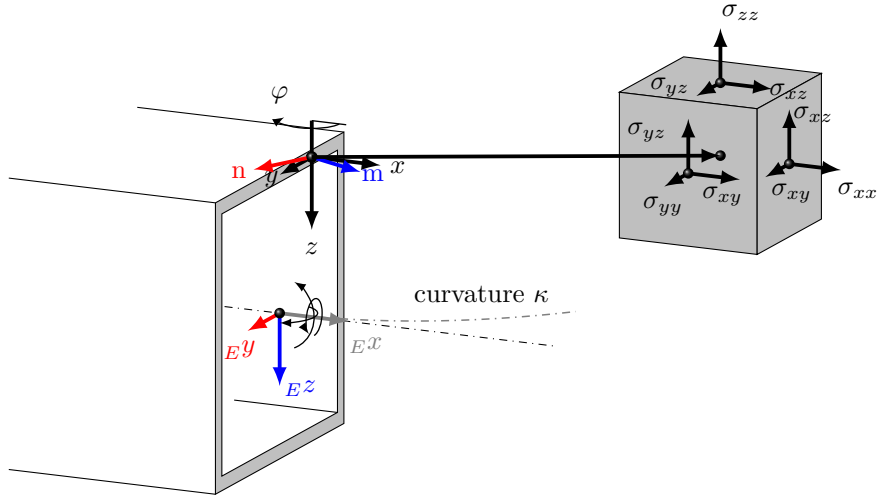

Since a beam generally bends and twists under load, this results in a multi-axial stress state with normal and shear components. The stress state at a material point $\leftidx{_E}{\mathbf{r}}{_p}$ (\ref{eq:materialPoint}) is described by the symmetric stress tensor
\begin{equation}
\bm{\Sigma}_{\text{3D}}=
\begin{bmatrix}
\sigma_{xx} & \sigma_{xy} & \sigma_{xz}\\
\sigma_{xy} & \sigma_{yy} & \sigma_{yz}\\
\sigma_{xz} & \sigma_{yz} & \sigma_{zz}
\end{bmatrix}
\end{equation}
which leads through $\bm{\sigma}_n=\bm{\Sigma}\mathbf{n}$ to the stress vector, which depends on the cutting plane with normal vector $\mathbf{n}$. Since the components under consideration are beams with thin walled cross sections, the simplification to a plane stress state, characterized by $\sigma_{zz}=0$, $\sigma_{xz}=0$ and $\sigma_{yz}=0$ , is permissible. Therefore the stress tensor is then
\begin{equation}
\bm{\Sigma}_{\text{2D}}=
\begin{bmatrix}
\sigma_{xx} & \sigma_{xy} & 0\\
\sigma_{xy} & \sigma_{yy} & 0\\
0 & 0 & 0
\end{bmatrix}.
\end{equation}
Introducing the cutting frame $\mathcal{F}_c$, under the cutting angle $\varphi$ with the axes $\mathbf{n}$ and $\mathbf{m}$, see Fig. \ref{fig:stressStateBeam}, the components of the stress vector are
\begin{equation}
\label{eq:cuttingStressNormal}
\sigma_{nn}=\frac{\sigma_{xx}+\sigma_{yy}}{2}+\frac{\sigma_{xx}-\sigma_{yy}}{2}\cos{\left(2\varphi\right)}+\sigma_{xy}\sin{\left(2\varphi\right)}
\end{equation}
and
\begin{equation}
\label{eq:cuttingStressShear}
\sigma_{nm}=-\frac{\sigma_{xx}-\sigma_{yy}}{2}\sin{\left(2\varphi\right)}+\sigma_{xy}\cos{\left(2\varphi\right)}.
\end{equation}

Simulation of the elastic beam model from section \ref{sec:modeling} according to the trajectory and control in section \ref{sec:loadCases} provides the time histories $v(\xi,t)$, $w(\xi,t)$ and $\vartheta(\xi,t)$ of the displacements of the beam axis. The resulting normal and shear stresses
are calculated by (\ref{eq:multiaxialHooke}). For every cutting angle $\varphi$ the shear stress
\begin{equation}
\label{eq:shearStressPhi}
\tau_{\varphi}(\xi,t):=\sigma_{nm}(\xi,t)=-\frac{1}{2}\sigma_{xx}(\xi,t)\sin{\left(2\varphi\right)}+\sigma_{xy}(\xi,t)\cos{\left(2\varphi\right)}
\end{equation}
is calculated by (\ref{eq:cuttingStressShear}).\\

Material characteristics are determined experimentally for uniaxial loading. Therefore, a hypothesis is required to relate the multiaxial stress state to a load-equivalent uniaxial stress state. Tresca's hypothesis is such a relation \cite{CunhaTresca,lee2011metal}. According to this hypothesis the equivalent stress is
\begin{equation}
\label{eq:Tresca}
\sigma_{\text{Tresca}}=2\tau_{\varphi}(\xi,t),
\end{equation}
which is assumed to be responsible for the failure of the component. It is mainly used for tough materials under static stress, but is also used for alternating stresses, as it tends to provide a higher equivalent stress and, as a result, greater safety. This hypothesis is used in the subsequent procedure.\\

\subsection{Identifying Damage Events using Rainflow-Counting}
As the equivalent stress (\ref{eq:Tresca}) varies over time the material is loaded and unloaded. A load-cycle with mean stress $\sigma_m$ and amplitude stress $\sigma_a$ is generally defined by a closed loop of the material's uniaxial stress-strain response. Every load cycle, as long as it reaches a certain stress level, see subsection \ref{subsec:damageGrowth}, further damages the material. In order to be able to accumulate the damage, the damage events must be extracted from the load sequence calculated from (\ref{eq:shearStressPhi}) by (\ref{eq:Tresca}). Identifying a load cycle, as depicted in Fig. \ref{fig:loadCycleUniform}, is not straightforward for an arbitrary load history Fig. \ref{fig:loadCycleArbitrary}.\\

\pgfplotsset{
compat=newest,
/pgfplots/myylabel absolute/.style={%
  /pgfplots/every axis y label/.style={at={(0,0.5)},xshift=#1,rotate=90},
  /pgfplots/every y tick scale label/.style={
    at={(0,1)},above right,inner sep=0pt,yshift=0.3em
   }
  }
}

\definecolor{mycolor}{rgb}{0.23935,0.30085,0.54084}%

\definecolor{c1}{rgb}{0,      0.4470,	0.7410}%
\definecolor{c2}{rgb}{0.8500, 0.3250,	0.0980}%
\definecolor{c3}{rgb}{0.9290, 0.6940,	0.1250}%
\definecolor{c4}{rgb}{0.4940, 0.1840,	0.5560}%
\definecolor{c5}{rgb}{0.4660, 0.6740,	0.1880}%
\definecolor{c6}{rgb}{0.6350, 0.0780,	0.1840}%

\def\myWidth{8cm}
\def\myHeight{1.3cm}

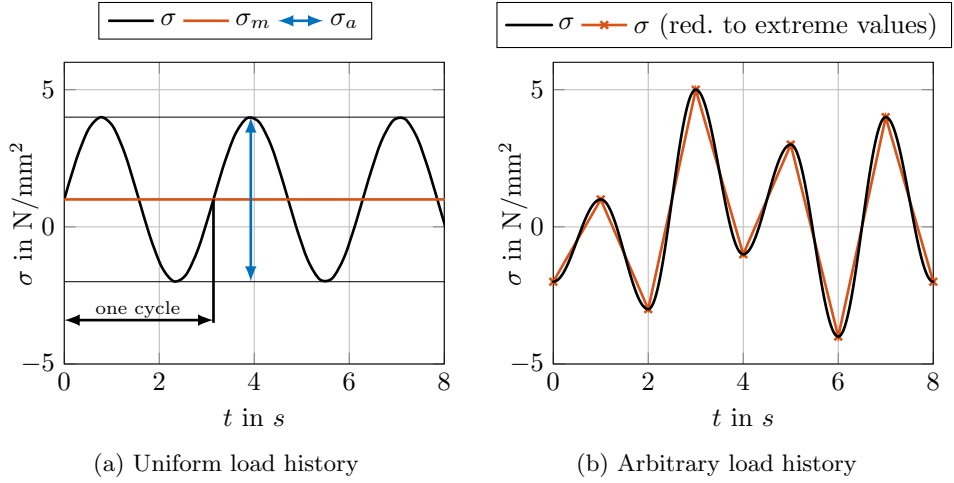
\begin{figure}[ht]
\centering
\subfloat[Uniform load history]{
\begin{tikzpicture}
\begin{axis}[
	name = plot1,
	width= 5cm,
	height= 4cm,
	scale only axis,
	xmin=0,
	xmax=8,
	ymin=-5,
	ymax=6,
	xlabel style={align=center,font=\color{white!15!black}},
	xlabel={$t$ in $s$},
	ylabel style={align=center,font=\color{white!15!black}},
	ylabel={$\sigma$ in $\si{N/mm^2}$},
	myylabel absolute=-15pt,
	axis background/.style={fill=white},
	xmajorgrids,
	ymajorgrids,
	xlabel style={font=\normalsize},
	ylabel style={align=center,font=\normalsize},
	ticklabel style={font=\normalsize}, set layers,
	legend style={
		at={(0.45,1.2)},
		anchor=north,
		font=\normalsize,
	legend columns=3,
	legend transposed=false},
	]
	\addplot[domain=0:4*pi, samples=50, smooth, color=black, line width=1pt] {1+3*sin(deg(2*x))};
	\addlegendentry{$\sigma$};
	\addplot[mark=none, color = c2, line width=1pt] coordinates {(0,1) (4*pi,1)};
	\addlegendentry{$\sigma_m$};
	\addplot[latex-latex, mark=none, color = c1, line width=1pt] coordinates {(1.25*pi,-2) (1.25*pi,4)};
	\addlegendentry{$\sigma_a$};
	\addplot[mark=none, black, line width=1pt] coordinates {(pi,1) (pi,-3.5)};
	\addplot[latex-latex, mark=none, black, line width=1pt] coordinates {(0,-3.4) (pi,-3.4)};
	\node [] at (1.55,-3) {\footnotesize one cycle};
	\addplot[mark=none, color = black, line width=0.1mm] coordinates {(0,4) (4*pi,4)};
	\addplot[mark=none, color = black, line width=0.1mm] coordinates {(0,-2) (4*pi,-2)};
	\label{fig:loadCycleUniform}
\end{axis}
\end{tikzpicture}
}\hfil
\subfloat[Arbitrary load history]{
\begin{tikzpicture}
\begin{axis}[
	name = plot2,
	width= 5cm,
	height= 4cm,
	scale only axis,
	xmin=0,
	xmax=8,
	ymin=-5,
	ymax=6,
	xlabel style={align=center,font=\color{white!15!black}},
	xlabel={$t$ in $s$},
	ylabel style={align=center,font=\color{white!15!black}},
	ylabel={$\sigma$ in $\si{N/mm^2}$},
	myylabel absolute=-15pt,
	axis background/.style={fill=white},
	xmajorgrids,
	ymajorgrids,
	xlabel style={font=\normalsize},
	ylabel style={align=center,font=\normalsize},
	ticklabel style={font=\normalsize}, set layers,
	legend style={
		at={(0.45,1.2)},
		anchor=north,
		font=\normalsize,
	legend columns=3,
	legend transposed=false},
	]
	\addplot+[solid, color=black, line cap=round, mark=none, line width=1pt, on layer=axis foreground] table[x=Time, y=Stress, col sep=semicolon] {data.csv};
	\addlegendentry{$\sigma$};
	\addplot+[color=c2, line width=1pt, mark=x] coordinates {(0,-2) (1,1) (2,-3) (3,5) (4,-1) (5,3) (6,-4) (7,4) (8,-2)};
	\addlegendentry{$\sigma$ (red. to extreme values)};
	\label{fig:loadCycleArbitrary}
\end{axis}
\end{tikzpicture}
}
\caption{Definition of mean and amplitude stress (left) and exemplary load history (right)}
\label{fig:loadCycle}
\end{figure}

\begin{figure}[ht]
\begin{center}
\begin{tikzpicture}[scale = 0.75]

\coordinate (origin1) at (0,0);
\coordinate (xPlus1) at ($(origin1)+(3,0)$);
\coordinate (xMinus1) at ($(origin1)+(-3,0)$);
\coordinate (yPlus1) at ($(origin1)+(0,-9)$);
\draw[-latex, line width = 0.25mm] (xMinus1) -- (xPlus1);
\draw[-latex, line width = 0.25mm] ($(origin1)+(0,0.2)$) -- (yPlus1);
\node[] at ($(xPlus1)+(0,0.5)$) {$\sigma$};
\node[] at ($(yPlus1)+(-0.5,0)$) {$t$};

\draw [line width = 0.5mm] (-1,0) -- (0.5,-1) -- (-1.5,-2) -- (2.5,-3) -- (-0.5,-4) -- (1.5,-5) -- (-2,-6) -- (2,-7) -- (-1,-8);
\draw [line width = 0.25mm, red] (-1,0) -- (0.5,-1) -- (0.5,-2);
\draw [line width = 0.25mm, red] (-1.5,-2) -- (2.5,-3) -- (2.5,-6);
\draw [line width = 0.25mm, red]  (-0.5,-4) -- (1.5,-5) -- (1.5,-6);
\draw [line width = 0.25mm, red]  (-2,-6) -- (2,-7) -- (2,-9);

\draw [line width = 0.25] (-1.5,-2) -- (-1.5,-3.5);
\draw [line width = 0.25] (2.5,-3) -- (2.5,-3.5);
\draw [latex-latex, line width = 0.25] (-1.5,-3.4) -- (2.5,-3.4);
\filldraw (0.5,-3.4) circle (2pt);

\node [] at (-0.5,-3.2) {$\sigma_a$};
\node [] at (0.5,-3) {$\sigma_m$};

\node [] at (-1.5,0) {$\bigtriangledown$};
\node [] at (-2,-2) {$\bigtriangledown$};
\node [] at (-1,-4) {$\bigtriangledown$};
\node [] at (-2.5,-6) {$\bigtriangledown$};

\coordinate (origin2) at (7,0);
\coordinate (xPlus2) at ($(origin2)+(3,0)$);
\coordinate (xMinus2) at ($(origin2)+(-3,0)$);
\coordinate (yPlus2) at ($(origin2)+(0,-9)$);
\draw[-latex, line width = 0.25mm] (xMinus2) -- (xPlus2);
\draw[-latex, line width = 0.25mm] ($(origin2)+(0,0.2)$) -- (yPlus2);
\node[] at ($(xPlus2)+(0,0.5)$) {$\sigma$};
\node[] at ($(yPlus2)+(-0.5,0)$) {$t$};

\draw [line width = 0.5mm] (-1+7,0) -- (0.5+7,-1) -- (-1.5+7,-2) -- (2.5+7,-3) -- (-0.5+7,-4) -- (1.5+7,-5) -- (-2+7,-6) -- (2+7,-7) -- (-1+7,-8);

\draw [line width = 0.25mm, orange] (7.5,-1) -- (5.5,-2) -- (5.5,-3);
\draw [line width = 0.25mm, blue] (9.5,-3) -- (6.5,-4) -- (6.5,-5.57) -- (5,-6) -- (5,-7);
\draw [line width = 0.25mm, green] (8.5,-5) -- (6.6,-5.54);
\draw [line width = 0.25mm, blue] (9,-7) -- (6,-8) -- (6,-9);

\draw [line width = 0.1mm, gray] (-2,0.1) -- (-2,-8.5);
\draw [line width = 0.1mm, gray] (-1,0.1) -- (-1,-8.5);
\draw [line width = 0.1mm, gray] (1,0.1) -- (1,-8.5);
\draw [line width = 0.1mm, gray] (2,0.1) -- (2,-8.5);

\draw [line width = 0.1mm, gray] (-2+7,0.1) -- (-2+7,-8.5);
\draw [line width = 0.1mm, gray] (-1+7,0.1) -- (-1+7,-8.5);
\draw [line width = 0.1mm, gray] (1+7,0.1) -- (1+7,-8.5);
\draw [line width = 0.1mm, gray] (2+7,0.1) -- (2+7,-8.5);

\node [rotate = -90] at (-2,0.5) {-4};
\node [rotate = -90] at (-1,0.5) {-2};
\node [rotate = -90] at (0,0.5) {0};
\node [rotate = -90] at (1,0.5) {2};
\node [rotate = -90] at (2,0.5) {4};

\node [rotate = -90] at (-2+7,0.5) {-4};
\node [rotate = -90] at (-1+7,0.5) {-2};
\node [rotate = -90] at (0+7,0.5) {0};
\node [rotate = -90] at (1+7,0.5) {2};
\node [rotate = -90] at (2+7,0.5) {4};

\node[shape=circle, draw, inner sep=2pt] at (5.5,-8.5) {a};
\node[shape=circle, draw, inner sep=2pt] at (7,-4.8) {b};
\node[shape=circle, draw, inner sep=2pt] at (6,-3) {c};

\node [] at (8,-1) {$\bigtriangleup$};
\node [] at (10,-3) {$\bigtriangleup$};
\node [] at (9,-5) {$\bigtriangleup$};
\node [] at (9.5,-7) {$\bigtriangleup$};

\end{tikzpicture}
\caption{Pagoda roof method for compressive valleys (left) and tensile peaks (right)}
\label{fig:PagodaRoof}
\end{center}
\end{figure}
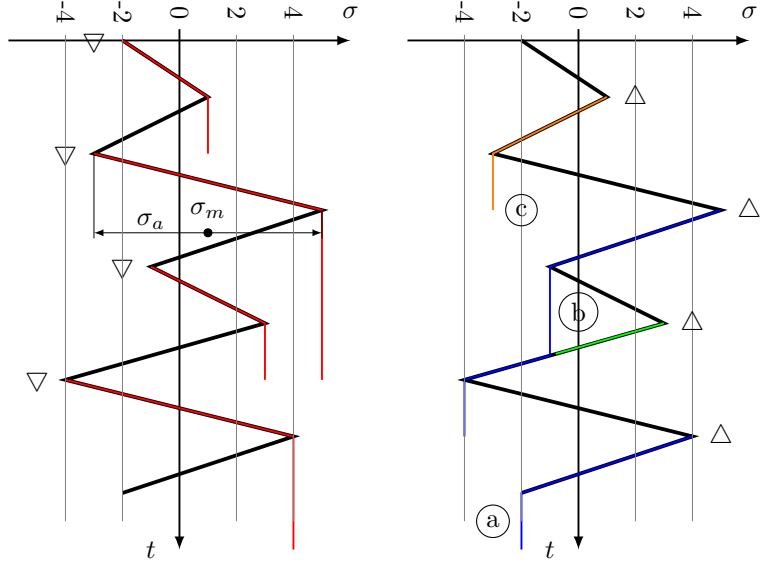

The rainflow-counting algorithm was developed and proposed by Matsuichi and Endo in 1968 \cite{Matsuishi1968}. The continuous time history $\sigma(t)$ of the equivalent stress (\ref{eq:Tresca}) is reduced to a set of tensile peaks $\bigtriangleup$ and compressive valleys $\bigtriangledown$, see Fig. \ref{fig:PagodaRoof}. The following steps are then carried out:
\begin{enumerate}
\item Every tensile peak  (Fig. \ref{fig:PagodaRoof} right) / compressive valley  (Fig. \ref{fig:PagodaRoof} left) is the source of a flow of water, that would drip down the roofs of a pagoda. Such a flow terminates if one of the following conditions holds:
\begin{itemize}
\item[a.)] The flow reaches the end of the load duration,\\
\item[b.)] it merges with a flow that started before (e.\,g. Fig. \ref{fig:PagodaRoof} right where green merges into blue),\\
\item[c.)] the next tensile peak is of greater or equal magnitude (e.\,g. Fig. \ref{fig:PagodaRoof} right where orange terminates because blue has a higher starting stress).
\end{itemize}
\item Each flow, whether tensile or compressive, represents a half cycle with stress magnitude equal to the difference between the stresses at its starting and ending points.\\
\item  One tensile and compressive half cycle each with equal magnitudes are paired up to full cycles. Some half cycles might remain after this step.
\end{enumerate}

The version of this algorithm, used in the subsequent example, is implemented according to the ASTM E 1049 standard \cite{ASTM}.\\

The mean- and amplitude stresses $\sigma_m$ and $\sigma_a$ are naturally continuous quantities. The expected stress ranges are divided into $N_m$ and $N_a$ subranges respectively, which grids the $\sigma_m$-$\sigma_a$ domain under consideration into $N_mN_a$ clusters also referred to as load collectives and referenced by their respective center points $(\sigma_m^{\ast},\sigma_a^{\ast})$. A load collective is a collection of load cycles whose mean and amplitude stresses $(\sigma_m,\sigma_a)$ lie within the same cluster $(\sigma_m^{\ast},\sigma_a^{\ast})$. Counting the number $N(\sigma_m^{\ast},\sigma_a^{\ast})$ of occurring events for every cluster results in a histogram, also referred to as rainflow matrix  $\mathbf{R}$.

\subsection{Damage Growth $\Delta D$}\label{subsec:damageGrowth}
\begin{figure}[ht]
\begin{center}
\begin{tikzpicture}[]

\draw[-latex, line width = 0.25mm] (0,0) -- (4,0);
\draw[-latex, line width = 0.25mm] (0,0) -- (0,4);
\node[] at (4.2,-0.5) {$\sigma_m$};
\node[] at (-0.5,4) {$\sigma_a$};
\draw[line width = 0.3mm] (0,3) -- (2,2.5) -- (3.5,0);
\draw[-latex, line width = 0.25mm] (1,0) -- (1,2.75);
\draw[-latex, line width = 0.25mm] (1,2.75) -- (0,2.75);
\node[] at (1,-0.5) {$\sigma_m^*$};
\node[] at (-0.5,2.7) {$\sigma_{D,a}^*$};
\node[] at (-0.5,3.05) {$\sigma_w$};
\node[] at (3.5,-0.5) {$R_e$};

\draw[-latex, line width = 0.25mm] (5,0) -- (9,0);
\draw[-latex, line width = 0.25mm] (5,0) -- (5,4);
\node[] at (9.2,-0.5) {$N$};
\node[] at (4.5,4) {$\sigma_a$};
\draw[line width = 0.3mm] (5,3) -- (6.5,3) -- (7.5,0.75) -- (8.5,0.75);
\draw[line width = 0.15mm] (5,0.75) -- (7.5,0.75);
\draw[line width = 0.15mm] (6.5,0) -- (6.5,3);
\draw[line width = 0.15mm] (7.5,0) -- (7.5,0.75);
\draw[-latex, line width = 0.25mm] (5,1.5) -- (7.17,1.5);
\draw[-latex, line width = 0.25mm] (7.17,1.5) -- (7.17,0);
\node[] at (6.5,-0.2) {$2.10^4$};
\node[] at (7.17,-0.75) {$N(\sigma_a^*)$};
\node[] at (7.5,-0.2) {$2.10^6$};
\node[] at (4.5,0.75) {$\sigma_{D,a}^*$};
\node[] at (4.5,1.5) {$\sigma_a^*$};
\node[] at (4.5,3) {$R_e$};

\node[shape=circle, draw, inner sep=2pt] at (1,-1.1) {1};
\node[shape=circle, draw, inner sep=2pt] at (-1.25,2.7) {2};
\node[shape=circle, draw, inner sep=2pt] at (8,1.25) {3};
\node[shape=circle, draw, inner sep=2pt] at (3.75,1.5) {4};
\node[shape=circle, draw, inner sep=2pt] at (7.17,-1.35) {5};

\end{tikzpicture}
\caption{Haigh-diagram (left) and Wöhler-diagram (right)}
\label{fig:HaighWoehler}
\end{center}
\end{figure}
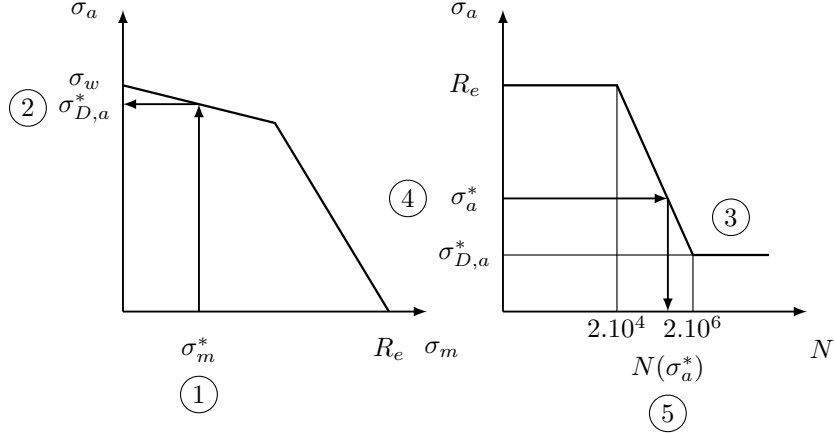

A damage increase is assigned to each load collective $(\sigma_m^{\ast},\sigma_a^{\ast})\in\mathbf{R}$. This is calculated from the number of cycles $N(\sigma_m^{\ast},\sigma_a^{\ast})$ and the number of fatigue-resistant load cycles in this stress range to
\begin{equation}
\Delta D=\frac{N(\sigma_m^{\ast},\sigma_a^{\ast})}{N(\sigma_{D,a}^{\ast})}.
\end{equation}
The number of fatigue-resistant load cycles $N(\sigma_{D,a}^{\ast})$ (step 5, see Fig. \ref{fig:HaighWoehler}) is extracted from the Wöhler \cite{MURAKAMI2021106138} diagram Fig. \ref{fig:HaighWoehler} (right), which is synthetically computed (with the characteristic point of step 3) according to
\begin{equation}
\frac{\log{\left(R_e\right)}-\log{\left(\sigma_{D,a}^{\ast}\right)}}{\log{\left(2\cdot10^6\right)}-\log{\left(2\cdot10^4\right)}}=\frac{\log{\left(R_e\right)}-\log{\left(\sigma_a^{\ast}\right)}}{\log{\left(N(\sigma_a^{\ast})\right)}-\log{\left(2\cdot10^4\right)}},
\end{equation}
for the specific fatigue-resistant amplitude stress  $\sigma_{D,a}^{\ast}$ (step 4). This (step 2) in turn is obtained by inserting the load cycle specific mean stress (step 1) into the fatigue strength diagram according to Haigh \cite{SENDECKYJ2001347}, see Fig. \ref{fig:HaighWoehler} (left). The yield strength is denoted by $R_e$. 

\subsection{Damage Accumulation}
The damage hypothesis is the way in which the total damage is calculated from the damage increments. When calculating those increments in subsection \ref{subsec:damageGrowth}, the implicit assumption is already made that the chronological order of occurrence of the individual load cycles is irrelevant. Particular materials show hardening effects due to load, which would mean that the sequence of the loads plays a crucial role. In this paper, the linear damage accumulation
\begin{equation}
D=\sum\limits_i\Delta D_i,
\end{equation}
according to Palmgren and Miner \cite{Miner} is used, which neglects those effects. The overall damage $D$ is calculated for every cutting angle and the highest damage is used. This approach is referred to as \textit{method of the critical cutting plane} \cite{Fatemi2011}.\\

The methodology is summarized in Alg. \ref{alg:lifetimeEstimation} and Alg. \ref{alg:lifetimeEstimationSubalgorithm}.

\begin{algorithm}[ht]
\caption{Lifetime estimation of a critical point in the robot's structure on basis of the \textit{method of the critical cutting plane}.}\label{alg:lifetimeEstimation}
\KwIn{set of cutting angles $\Phi$}
\KwData{normal and shear stresses $\sigma_{xx}(\xi,t)$ and $\sigma_{xy}(\xi,t)$ at $\xi=\xi_{\text{crit.}}$}
\For{$\varphi\in\Phi$}{
- calculate shear stress (\ref{eq:shearStressPhi}) and equivalent stress (\ref{eq:Tresca})\;
- calculate rainflow matrix $\mathbf{R}$\;
\For{$\mathbf{R}$}{
- call Alg. \ref{alg:lifetimeEstimationSubalgorithm}\;
- $D(\varphi):=D(\varphi)+\Delta D$ (Palmgren-Miner)\;
}
$D_{\text{max}}:=\text{max}\left(D(\varphi),D_{\text{max}}\right)$ (Critical Cutting Plane)
}
\KwResult{estimated lifetime $t_{\text{life}}:=t_{\text{task}}/D_{\text{max}}$}
\end{algorithm}

\begin{algorithm}[ht]
\caption{Subalgorithm to assign a damage increment to each load spectrum of the rainflow matrix $\mathbf{R}$.}\label{alg:lifetimeEstimationSubalgorithm}
\KwIn{material specific Haigh diagram}
\KwData{$\left(\sigma_m^{\ast},\sigma_a^{\ast}\right)\in\mathbf{R}$}
\Begin(calculate the related damage growth $\Delta D$){
- fatigue strength $\sigma_{D,a}^{\ast}$ for $\sigma_m^{\ast}$ from Haigh diagram\;
- $N\left(\sigma_a^{\ast}\right)$ from Wöhler line for $\sigma_m^{\ast}$
}		
\KwResult{$\Delta D=N\left(\sigma_m^{\ast},\sigma_a^{\ast}\right)/N\left(\sigma_a^{\ast}\right)$ associated to load spectrum}
\end{algorithm}

\section{Example}\label{applicationAndResults}
The considered articulated robot has three degrees of freedom $\mathbf{q}^T=(q_1\,q_2\,q_3)$, see Fig. \ref{fig:configMatandRob} (right), with gearbox- and link-elasticities. To keep the example clear and concise, only the wall thicknesses of the links are varied. The reference robot's links have a quadratic cross section with edge length $a=35\,\si{mm}$ and wall thickness $t_i=4\,\si{mm},\,i=1,2$. Varying $t_1$ (first link) and $t_2$ (second link) between $1\,\si{mm}$ and $6\,\si{mm}$ leads to $36$ possible candidate designs. As only two parameters are varied, the set of configurations $\mathbf{C}$ can be written down as shown in Fig. \ref{fig:configMatandRob}. The reference is configuration $22$. The numbering corresponds to that in Fig. \ref{fig:paretoDiag} and Fig. \ref{fig:lifetimeDiag}.\\

\subsection{Parameter Study and Determination of Candidate Designs}
The intended task for this example is picking an object from a position with correspoing robot configuration $\mathbf{q}_{\text{pick}}$ and transporting it to a goal position with $\mathbf{q}_{\text{place}}$. For every $q_i$, a trajectory with a trapezoidal acceleration profile is planned under consideration of velocity-, accerleration- and jerk-limits.

\begin{figure}[ht]
\begin{center}
\begin{tikzpicture}[
declare function={
	c(\i,\j)=(\j-1)*6+\i;
}
]
\foreach \i in {1,...,6}
{
    \draw[line width=0.1mm] (\i/2,0) -- (\i/2,7/2);
    \draw[line width=0.1mm] (0,\i/2) -- (7/2,\i/2);
\node[] at (\i/2+0.25,0.25) {\i};
\node[] at (0.25,\i/2+0.25) {\i};
}

\draw[line width=0.5mm] (7/2,1/2) -- (1/2,1/2) -- (1/2,7/2);

\foreach \i in {1,...,6}
{
\draw[red,line width=0.5mm] (\i/2-1/2,\i/2) -- (\i/2,\i/2) -- (\i/2,\i/2+1/2);
}

\foreach \i in {1,...,6}
{
	\foreach \j in {1,...,6}
	{
		\tikzmath{\v=int((\j-1)*6+\i);}
		\node[] at (\i/2+0.25,\j/2+0.25) {$\v$};
	}
}

\node[align = left] at (1.75,-0.5) {$t_1$ in mm};
\node[align = left, rotate=90] at (-0.5,1.75) {$t_2$ in mm};

\coordinate (I) at (6,-1);
\draw[line width = 1mm] ($(I)+(-0.5,0)$) -- ($(I)+(0.5,0)$);
\draw[line width = 1mm] (I) -- ($(I)+(0,1)$);

\draw[line width = 1mm] ($(I)+(-0.3,1)$) -- ($(I)+(0.3,1)$);
\draw[line width = 1mm] ($(I)+(-0.3,1.2)$) -- ($(I)+(0.3,1.2)$);

\draw[line width = 1mm] ($(I)+(0,1.2)$) -- ++(0,1) -- ++(55:3.5) -- ++(-30:3.5);
\coordinate (j1) at ($(I)+(0,1.1)$);
\coordinate (j2) at ($(I)+(0,2.1)$);
\coordinate (j3) at ($(j2)+(55:3.5)$);
\coordinate (l1) at ($(j2)+(55:1.75)$);
\coordinate (l2) at ($(j3)+(-30:1.75)$);
\draw[fill] (j2) circle (1.5mm);
\draw[fill] (j3) circle (1.5mm);
\node[right] at ($(j1)+(0.3,0)$) {$q_1$};
\node [left] at ($(j2)-(0.2,0)$) {$q_2$};
\node [above] at ($(j3)+(0,0.2)$) {$q_3$};

\node[rotate=55, above = 0.25cm] at (l1) {first link - $t_1$};
\node[rotate=-30, above = 0.25cm] at (l2) {second link - $t_2$};

\end{tikzpicture}
\caption{Configurations for the chosen parameter ranges (left) and scheme of the articulated 3DOF elastic link robot (right)}
\label{fig:configMatandRob}
\end{center}
\end{figure}
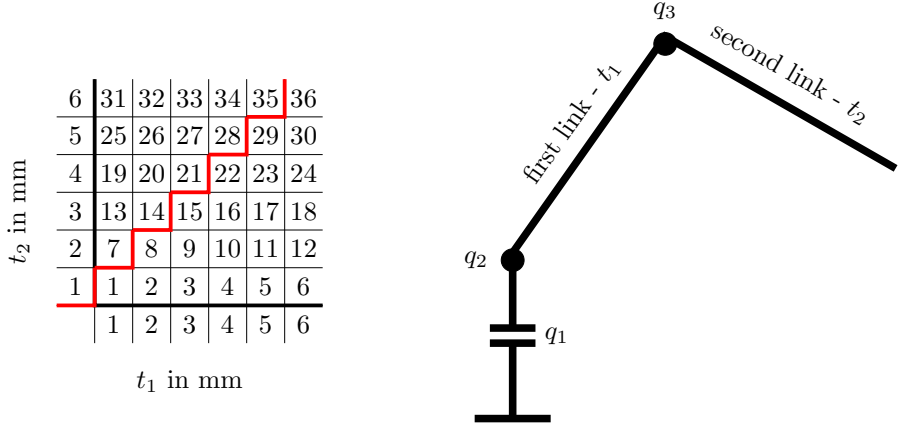

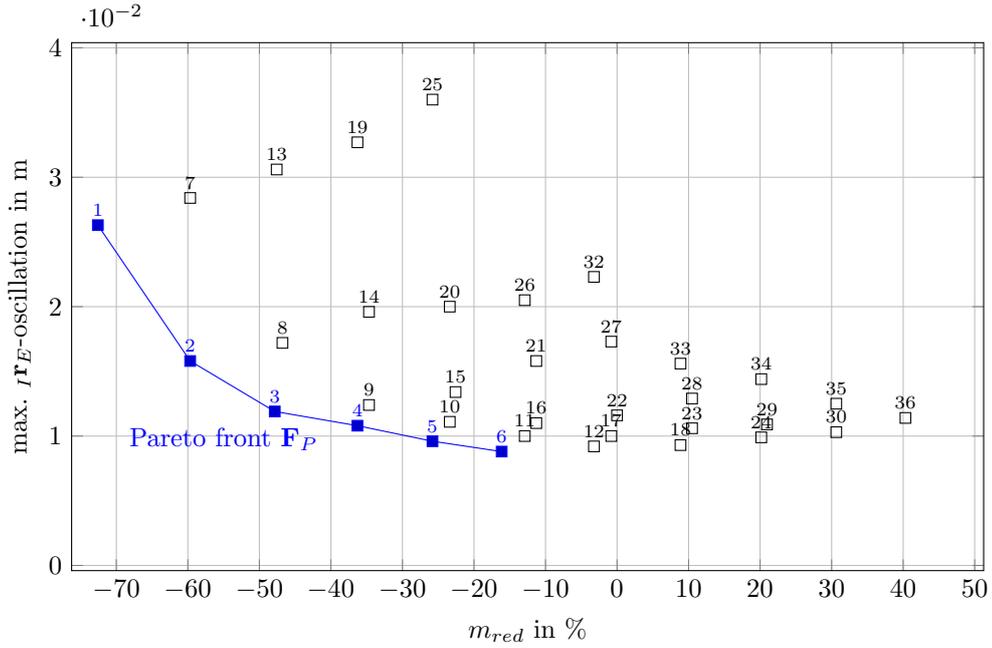
\begin{figure}[ht]
\begin{center}
\begin{tikzpicture}
\begin{axis}[
	name = plot1,
	width= 12cm,
	height= 7cm,
	scale only axis,
	xmin=-75,
	xmax=50,
	ymin=0,
	ymax=0.04,
	xlabel style={align=center,font=\color{white!15!black}},
	xlabel={$m_{red}$ in \%},
	ylabel style={align=center,font=\color{white!15!black}},
	ylabel={max. $\leftidx{_I}{\mathbf{r}}{_E}$-oscillation in m},
	axis background/.style={fill=white},
	xmajorgrids,
	ymajorgrids,
	xlabel style={font=\normalsize},
	ylabel style={align=center,font=\normalsize},
	ticklabel style={font=\normalsize}, set layers,
	legend style={
		at={(0.45,1.5)},
		anchor=north,
		font=\small,
	legend columns=3,
	legend transposed=false},
nodes near coords,
enlargelimits=0.01
]

\addplot+ [
mark = square*,
point meta=explicit symbolic
] coordinates {
(-72.5806,0.0263) [\footnotesize1]
(-59.6774,0.0158) [\footnotesize2]
(-47.8506,0.0119) [\footnotesize3]
(-36.2903,0.0108) [\footnotesize4]
(-25.8065,0.0096) [\footnotesize5]
(-16.1290,0.0088) [\footnotesize6]
}; 
\addplot+ [black,
mark = square,
only marks,
point meta=explicit symbolic
] coordinates {
(-59.6774,0.0284) [\footnotesize7]
(-46.7742,0.0172) [\footnotesize8]
(-34.6774,0.0124) [\footnotesize9]
(-23.3871,0.0111) [\footnotesize10]
(-12.9032,0.0100) [\footnotesize11]
(-3.2258,0.0092) [\footnotesize12]
(-47.5806,0.0306) [\footnotesize13]
(-34.6774,0.0196) [\footnotesize14]
(-22.5806,0.0134) [\footnotesize15]
(-11.2903,0.0110) [\footnotesize16]
(-0.8065,0.0100) [\footnotesize17]
(8.8710,0.0093) [\footnotesize18]
(-36.2903,0.0327) [\footnotesize19]
(-23.3871,0.0200) [\footnotesize20]
(-11.2903,0.0158) [\footnotesize21]
(0,0.0116) [\footnotesize22]
(10.4839,0.0106) [\footnotesize23]
(20.1613,0.0099) [\footnotesize24]
(-25.8065,0.0360) [\footnotesize25]
(-12.9032,0.0205) [\footnotesize26]
(-0.8065,0.0173) [\footnotesize27]
(10.4839,0.0129) [\footnotesize28]
(20.9677,0.0109) [\footnotesize29]
(30.6452,0.0103) [\footnotesize30]
(-16.1290,0.0431) [\footnotesize31]
(-3.2258,0.0223) [\footnotesize32]
(8.8710,0.0156) [\footnotesize33]
(20.1613,0.0144) [\footnotesize34]
(30.6452,0.0125) [\footnotesize35]
(40.3226,0.0114) [\footnotesize36]
}; 
	
\end{axis}


\node [] at (2,1.75) {\textcolor{blue}{Pareto front $\mathbf{F}_P$}};

\end{tikzpicture}
\caption{Pareto diagram: max.$\leftidx{_I}{\mathbf{r}}{_E}$-oscillations via percentage mass reduction $m_{red}$ of the beams compared to the reference configuration}
\label{fig:paretoDiag}
\end{center}
\end{figure}

As can be seen in Fig. \ref{fig:paretoDiag} the configurations above the red line in Fig. \ref{fig:configMatandRob} lie within a region of the Pareto diagram which is quite uninteresting, also for the fact, that a second link thicker than the first one, does not make sense. Such configurations could be eliminated beforehand. One can identify groups of six configurations each (e.\,g. $1-7-13-19-25-31$), which lie on an imaginary curve. Those groups represent configurations with a constant thickness of the first elastic link. They get steeply rising as this thickness decreases and the first part of the kinematic chain is weakened. The weight reduction should be obtained further ahead in the kinematic chain or at least equivalently distributed. As expected, the oscillations of the endeffector increase, as the link stiffness decreases with ongoing mass reduction. With more solid, heavier arms and therefore a stiffer structure, the oscillations nevertheless increase again from a certain point. This is due to the gearbox elasticities, which are excited as the weight of the arms gets higher.\\

As one is interested in mass reduction, the Pareto front in Fig. \ref{fig:paretoDiag} defines a set $\hat{\mathbf{C}}=\{1,2,3,4,5,6\}$ of interesting configurations, which need to be further investigated.

\subsection{Fatigue Analysis of the Candidate Designs}
\begin{figure}[ht]
\begin{center}
\begin{tikzpicture}
\begin{axis}[
	name = plot1,
	width= 11cm,
	height= 5cm,
	scale only axis,
           scaled y ticks=base 10:-3,
	xmin=0,
	xmax=36,
	ymin=0,
	ymax=4000,
	xlabel style={align=center,font=\color{white!15!black}},
	xlabel={configuration \#},
	ylabel style={align=center,font=\color{white!15!black}},
	ylabel={lifetime in $h$},
	axis background/.style={fill=white},
	xmajorgrids,
	ymajorgrids,
	xlabel style={font=\normalsize},
	ylabel style={align=center,font=\normalsize},
	ticklabel style={font=\normalsize}, set layers,
	legend style={
		at={(0.45,1.5)},
		anchor=north,
		font=\small,
	legend columns=3,
	legend transposed=false},
nodes near coords,
enlargelimits=0.02
]

\addplot+ [black,
mark = square*,
only marks,
point meta=explicit symbolic
] coordinates {
(1,357.4835) [\footnotesize1]
(7,125.0560) [\footnotesize7]
(8,2856.2) [\footnotesize8]
(13,79.2570) [\footnotesize13]
(14,1126.3) [\footnotesize14]
(19,53.7651) [\footnotesize19]
(20,1045.5) [\footnotesize20]
(21,2956.8) [\footnotesize21]
(25,33.1010) [\footnotesize25]
(26,548.1084) [\footnotesize26]
(27,2956.8) [\footnotesize27]
(31,26.6737) [\footnotesize31]
(32,262.2964) [\footnotesize32]
(33,2856.2) [\footnotesize33]
(34,3061.6) [\footnotesize34]
}; 
\addplot+ [black,
mark=triangle*,
only marks,
point meta=explicit symbolic
] coordinates {
(2,3500) [\footnotesize2]
(3,3500) [\footnotesize3]
(4,3500) [\footnotesize4]
(5,3500) [\footnotesize5]
(6,3500) [\footnotesize6]
(9,3500) [\footnotesize9]
(10,3500) [\footnotesize10]
(11,3500) [\footnotesize11]
(12,3500) [\footnotesize12]
(15,3500) [\footnotesize15]
(16,3500) [\footnotesize16]
(17,3500) [\footnotesize17]
(18,3500) [\footnotesize18]
(22,3500) [\footnotesize22]
(23,3500) [\footnotesize23]
(24,3500) [\footnotesize24]
(28,3500) [\footnotesize28]
(29,3500) [\footnotesize29]
(30,3500) [\footnotesize30]
(35,3500) [\footnotesize35]
(36,3500) [\footnotesize36]
};   
	
\end{axis}

\end{tikzpicture}
\caption{Estimated lifetime in hours for every $c\in\mathbf{C}$}
\label{fig:lifetimeDiag}
\end{center}
\end{figure}
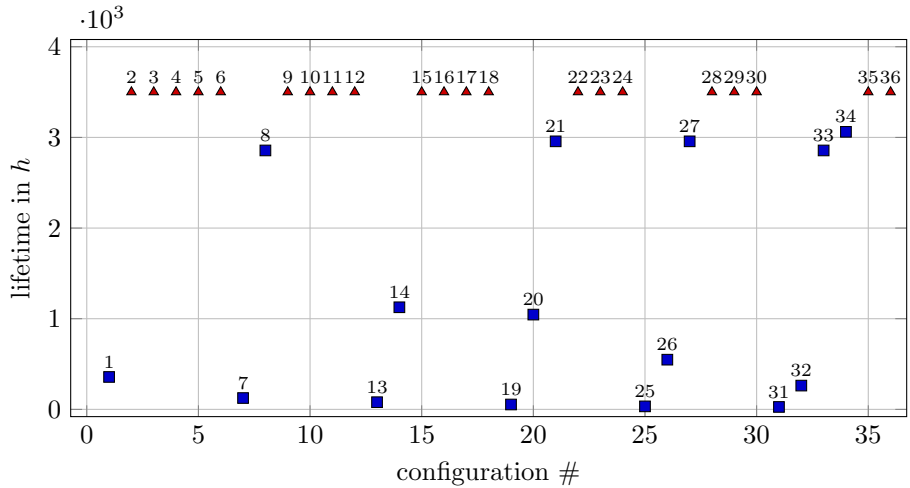

The dynamics simulations of the configurations $\hat{c}\in\hat{\mathbf{C}}\subset\mathbf{C}$ provide with the deformations of the elastic links and therefore the resulting stresses at a specific point. Applying Alg. \ref{alg:lifetimeEstimation} results in an estimated lifetime for each configuration, see Fig. \ref{fig:lifetimeDiag}. Some configurations do not reach stresses that go beyond the fatigue strength $\sigma_{D,a}^{\ast}$, e.\,g. $c=2$. Therefore our specific approach does not yield any damage to the structure which results in an (theoretically) infinite lifetime. The corresponding configurations are marked with an upright triangle. From Fig. \ref{fig:paretoDiag} and Fig. \ref{fig:lifetimeDiag}, one can conclude that configurations $\hat{\mathbf{C}}=\{2,3,4,5,6\}$ are the best with respect to mass reduction, endeffector oscillation and estimated lifetime.

\section{Conclusion and Outlook}\label{sec:conclusion}
A optimal design method is presented for optimizing a lumped parameter model of an elastic robot performing pick-and-place operations. The optimality criterion is a combination of the mass of moving links and vibration characteristics. An optimal design can be selected from a Pareto set. To his end, a method for lifetime estimation is presented. The latter combines the rainflow-counting algorithm and the critical cutting plane method using the Tresca equivalent stress hypothesis and assuming linear damage accumulation. The method is demonstrated for an elastic 3DOF robot.

The premise of this paper is that elastic deformations are primerily dues to the elastic links and that the link geometries are the only design parameters. It is hypothesized that further weight reduction is possible when elasticities in joints and gears and their mass are taken into account as optimization parameters. The latter becomed more important as the mass of joints and motors becomes more dominant when the links become slender. This will need additional input regarding the equivalent stiffness \cite{YinEtAl}.

\backmatter

\bmhead{Acknowledgments}
This work has been supported by the "LCM – K2 Center for Symbiotic Mechatronics" within the framework of the Austrian COMET-K2 program.





\bibliography{References_ZAUNER_Klaus_2024}

\end{document}